\documentclass{article}
\usepackage{ijcai19}

\usepackage{times}
\usepackage{soul}
\usepackage{url}
\usepackage[hidelinks]{hyperref}
\usepackage[utf8]{inputenc}
\usepackage[small]{caption}
\usepackage{graphicx}
\usepackage{amsmath}
\usepackage{booktabs}
\usepackage{tabularx}
\usepackage{xcolor}
\usepackage{amsmath}
\usepackage{amssymb}
\usepackage{amsfonts}
\usepackage[page]{appendix}
\usepackage{easyeqn}
\usepackage{algorithm}
\usepackage{algorithmicx}
\usepackage{algpseudocode}

\algrenewcommand\algorithmicrequire{\textbf{Input:}}
\algrenewcommand\algorithmicensure{\textbf{Output:}}

\def\RR{\mathbb{R}}
\def\EE{\mathbb{E}}
\def\xv{\mathbf{x}}

\def\wv{\mathbf{w}}

\def\kv{\mathbf{k}}

\newcommand{\T}{\mathrm{T}}

\def\eps{\epsilon}
\def\cond{\,|\,}

\def\pool{\mathcal{P}}

\title{Deeper Connections between Neural Networks and Gaussian Processes Speed-up Active Learning}

\author{
  Evgenii Tsymbalov\footnote{Contact Author}\and
  Sergei Makarychev\and
  Alexander Shapeev\And
  Maxim Panov\\
  \affiliations
  Skolkovo Institute of Science and Technology (Skoltech)
  \emails
  \{e.tsymbalov, sergei.makarychev, a.shapeev, m.panov\}@skoltech.ru
}

\begin{document}

\maketitle

\begin{abstract}
  Active learning methods for neural networks are usually based on greedy criteria which ultimately give a single new design point for the evaluation. Such an approach requires either some heuristics to sample a batch of design points at one active learning iteration, or retraining the neural network after adding each data point, which is computationally inefficient. Moreover, uncertainty estimates for neural networks sometimes are overconfident for the points lying far from the training sample.  In this work we propose to approximate Bayesian neural networks (BNN) by Gaussian processes, which allows us to update the uncertainty estimates of predictions efficiently without retraining the neural network, while avoiding overconfident uncertainty prediction for out-of-sample points. In a series of experiments on real-world data including large-scale problems of chemical and physical modeling, we show superiority of the proposed approach over the state-of-the-art methods.
\end{abstract}

\section{Introduction}
\label{sec:intro}
  In the modern applications of machine learning, especially in engineering design, physics, chemistry, molecular and materials modeling, the datasets are usually of limited size due to the expensive cost of computations. On the other side, such applications typically allow the calculation of additional data at points chosen by the experimentalist. Thus, there is the need to get the highest possible approximation quality with as few function evaluations as possible. \textit{Active learning} methods help to achieve this goal by trying to select the best candidates for further target function computation using the already existing data and machine learning models based on these data. 
  
  Modern active learning methods are usually based either on ensemble-based methods~\cite{Settles2012} or on probabilistic models such as Gaussian process regression~\cite{Sacks1989,Burnaev2015}. However, the models from these classes often don't give state-of-the-art results in the downstream tasks. For example, Gaussian process-based models have very high computational complexity, which in many cases prohibits their usage even for moderate-sized datasets. Moreover, GP training is usually done for stationary covariance functions, which is a very limiting assumption in many real-world scenarios. Thus, active learning methods applicable to the broader classes of models are needed.

  In this work, we aim to develop efficient active learning strategies for neural network models. Unlike Gaussian processes, neural networks are known for relatively good scalability with the dataset size and are very flexible in terms of dependencies they can model. However, active learning for neural networks is currently not very well developed, especially for tasks other than classification. Recently some uncertainty estimation and active learning approaches were introduced for Bayesian neural networks~\cite{Gal2017,Hafner2018}. However, still one of the main complications is in designing active learning approaches which allow for sampling points without retraining the neural network too often. Another problem is that neural networks due to their parametric structure sometimes give overconfident predictions in the areas of design space lying far from the training sample. Recently, \cite{Matthews2018}~proved that deep neural networks with random weights converge to Gaussian processes in the infinite layer width limit. Our approach aims to approximate trained Bayesian neural networks and show that obtained uncertainty estimates enable to improve active learning performance significantly.

  We summarize the main contributions of the paper as follows:
  \begin{enumerate}
      \item We propose to compute the approximation of trained Bayesian neural network with Gaussian process, which allows using the Gaussian process machinery to calculate uncertainty estimates for neural networks. We propose active learning strategy based on obtained uncertainty estimates which significantly speed-up the active learning with NNs by improving the quality of selected samples and decreasing the number of times the NN is retrained.
      
      \item The proposed framework shows significant improvement over competing approaches in the wide range of real-world problems, including the cutting edge applications in chemoinformatics and physics of oil recovery.
  \end{enumerate}
  In the next section, we discuss the problem statement and the proposed approach in detail.

\section{Active learning with Bayesian neural networks and beyond}
\label{sec:main}
\subsection{Problem statement}
  We consider a regression problem with an unknown function \(f(\xv)\) defined on a subset of Euclidean space \(\mathcal{X} \subset \RR^{d}\), where an approximation function \(\hat{f}(\xv)\) should be constructed based on noisy observations
  \begin{EQA}[c]
    y = f(\xv) + \eps
  \end{EQA} 
  with \(\eps\) being some random noise.
  
  We focus on \textit{active learning} scenario which allows to iteratively enrich training set by computing the target function in design points specified by the experimenter. More specifically, we assume that the initial training set \(D_{\rm{init}} = \{\xv_i, y_i = f(\xv_i) + \eps_i\}_{i = 1}^{N}\) with precomputed function values is given.
  %
  On top of it, we are given another set of points called ``pool''
    \(\pool = \{\xv_j \in \mathcal{X}\}_{j = 1}^{N_{\rm{pool}}}\),
  which represents unlabelled data. Each point \(\xv \in \mathcal{P}\) can be annotated by computing \(y = f(\xv) + \eps\) so that the pair \(\{\xv, y\}\) is added to the training set.
  
  The standard active learning approaches rely on the greedy point selection
  \begin{EQA}[c]
  \label{al_general}
    \xv_{new} = \arg \max_{\xv \in \pool} A(\xv \cond \hat{f}, D),
  \end{EQA}
  where \(A(\xv \cond \hat{f}, D)\) is so-called \textit{acquisition function}, which is usually constructed based on the current training set \(D\) and corresponding approximation function \(\hat{f}\).
  
  The most popular choice of acquisition function is the variance \(\hat{\sigma}^2(\xv \cond \hat{f}, D)\) of the prediction \(\hat{f}(\xv)\), which can be easily estimated in some machine learning models such as Random forest~\cite{Mentch2016} or Gaussian process regression~\cite{Rasmussen2004}. However, for many types of models (e.g., neural networks) the computation of prediction uncertainty becomes a nontrivial problem.

\subsection{Uncertainty estimation and active learning with Bayesian neural networks}
  In this work, we will stick to the Bayesian approach which treats neural networks as probabilistic models \(p(y \cond \xv, \wv)\). Vector of neural network weights \(\wv\) is assumed to be a random variable with some prior distribution \(p(\wv)\). The likelihood \(p(y \cond \xv, \wv)\) determines the distribution of network output at a point \(\xv\) given specific values of parameters \(\wv\).
  There is vast literature on training Bayesian networks (see~\cite{Graves2011} and~\cite{Paisley2012} among many others), which mostly targets the so-called variational approximation of the intractable posterior distribution \(p(\wv \cond D)\) by some easily computable distribution \(q(\wv)\).
 
  %
  The approximate posterior predictive distribution reads as:
  \begin{EQA}[c]
    q(y \cond \xv) = \int p(y \cond \xv, \wv) ~ q(\wv) ~ d \wv.
  \end{EQA}
  The simple way to generate random values from this distribution is to use the Monte-Carlo approach, which allows estimating the mean:
  \begin{EQA}[c]
    \EE_{q(y \cond \xv)} ~ y \approx \frac{1}{T} \sum_{t = 1}^T \hat{f}(\xv, \wv_t),
  \end{EQA}
  where the weight values \(\wv_t\) are i.i.d. random variables from distribution \(q(\wv)\). Similarly, one can use Monte-Carlo to estimate the approximate posterior variance \(\hat{\sigma}^2(\xv \cond \hat{f})\) of the prediction \(y\) at a point \(\xv\) and use it as an acquisition function:
  \begin{EQA}[c]
    A(\xv \cond \hat{f}, D) = \hat{\sigma}^2(\xv \cond \hat{f}).
  \end{EQA}
  We note that the considered acquisition function formally doesn't depend on the dataset \(D\) except for the fact that \(D\) was used for training the neural network \(\hat{f}\).

\begin{figure}[ht!]
    \hspace*{-0.5cm} 
    \centering
    \includegraphics[scale=.3]{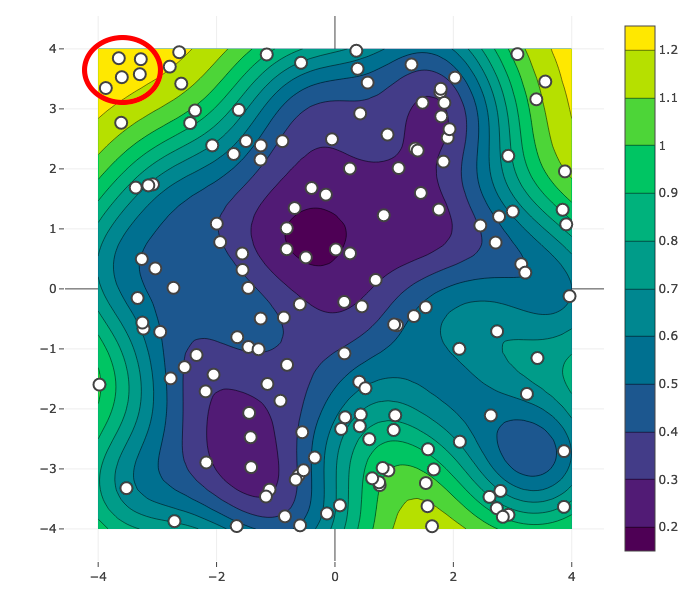} 
        \caption{Contour plot of neural network variance prediction for bivariate problem and pool points (white). Five points from the pool with maximum values of variance lie in the same region of design points (upper-left corner).}
    \label{fig:points}
  \end{figure}

  Let us note that the general greedy active learning approach~\eqref{al_general} by design gives one candidate point per active learning iteration. If one tries to use~\eqref{al_general} to obtain several samples with the same acquisition function, it usually results in obtaining several nearby points from the same region of design space, see Figure~\ref{fig:points}. Such behaviour is typically undesirable as nearby points are likely to have very similar information about the target function. Moreover, neural network uncertainty predictions are sometimes overconfident in out-of-sample regions of design space.

  There are several approaches to overcome these issues each having its drawbacks:
  \begin{enumerate}
    \item One may retrain the model after each point addition which may result in a significant change of the acquisition function and lead to the selection of a more diverse set of points. However, such an approach is usually very computationally expensive, especially for neural network-based models.

    \item One may try to add distance-based heuristic, which explicitly prohibits sampling points which are very close to each other and increase values of acquisition function for points positioned far from the training sample. Such an approach may give satisfactory results in some cases, however usually requires fine-tuning towards particular application (like the selection of specific distance function or choice the parameter value which determines whether two points are near or not), while its performance may degrade in high dimensional problems.

    \item One may treat specially normalized vector of acquisition function values at points from the pool as a probability distribution and sample the desired number of points based on their probabilities (the higher acquisition function value, the point is more likely to be selected). This approach usually improves over greedy baseline procedure. However, it still gives many nearby points.
  \end{enumerate}  
  In the next section, we propose the approach to deal with problems above by considering the Gaussian process approximation of the neural network.

\subsection{Gaussian process approximation of Bayesian neural network} 
  %
  Effectively, the random function
  \begin{EQA}[c]
    \hat{f}(\xv, \wv) = \EE_{p(y \cond \xv, \wv)} ~ y 
  \end{EQA}
  is the stochastic process indexed by \(\xv\).
  The covariance function of the process \(\hat{f}(\xv, \wv)\) is given by
  \begin{EQA}[c]
    k(\xv, \xv') = \EE_{q(\wv)} \bigl(\hat{f}(\xv, \wv) - m(\xv)\bigr) \bigl(\hat{f}(\xv', \wv) - m(\xv')\bigr),
  \end{EQA}
  where \(m(\xv) = \EE_{q(\wv)} \hat{f}(\xv, \wv)\).
  
\begin{figure}[ht!]
    \hspace*{-0.5cm} 
    \centering
    \includegraphics[scale=.07]{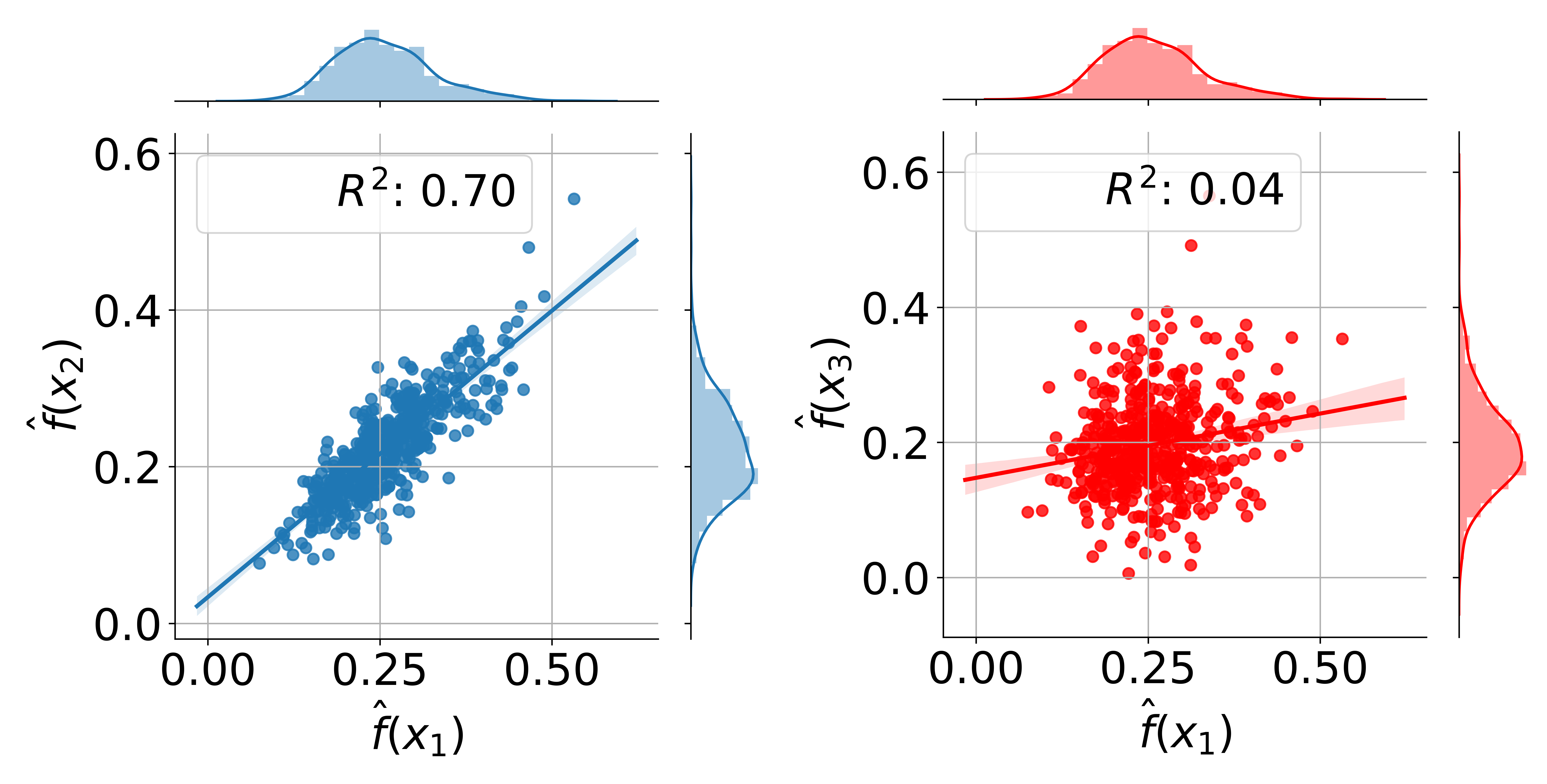} 
        \caption{Bivariate distribution plots for the stochastic NN output at points \(\xv_1, \xv_2\) and \(\xv_3\), where \(\xv_1\) is much closer to \(\xv_2\) in feature space than to \(\xv_3\). Both univariate and bivariate distributions are Gaussian-like, while the correlation between function values is much higher for closer points.}
    \label{fig:hists}
  \end{figure}

  As was shown in~\cite{Matthews2018,Lee2017} neural networks with random weights converge to Gaussian processes in the infinite layer width limit. However, one is not limited to asymptotic properties of purely random networks as Bayesian neural networks trained on real-world data exhibit near Gaussian behaviour, see the example on Figure~\ref{fig:hists}.

  We aim to make the Gaussian process approximation \(\hat{g}(\xv \cond \hat{f})\) of the stochastic process \(\hat{f}(\xv, \wv)\) and compute its posterior variance \(\hat{\sigma}^2(\xv \cond \hat{f}, X)\) given the set of anchor points \(X = \{\xv_i\}_{i = 1}^N\).
  Typically, \(X\) is a subset of the training sample.  Given \(X\), Monte-Carlo estimates \(\hat{k}(\xv', \xv'')\) of the covariance function \(k(\xv', \xv'')\) for every pair of points \(\xv', \xv'' \in X \cup \xv\) allow computing
  \begin{EQA}[c]
  \label{gp_approx_variance}
    \hat{\sigma}^2(\xv \cond \hat{f}, X) = \hat{k}(\xv, \xv) - \hat{\kv}^{\T}(\xv) \hat{K}^{-1} \hat{\kv}(\xv),
  \end{EQA}
  where \(\hat{K} = \bigl[\hat{k}(\xv_i, \xv_j)\bigr]_{i, j = 1}^{N}\) and \(\hat{\kv}(\xv) = \bigl(\hat{k}(\xv_1, \xv), \dots, \hat{k}(\xv_N, \xv)\bigr)^{\T}\).

  We note that only the trained neural network \(\hat{f}(\xv, \wv)\) and the ability to sample from the distribution \(q(\wv)\) are needed to compute \(\hat{\sigma}^2(\xv \cond \hat{f}, X)\).

\subsection{Active learning strategies}
  The benefits of the Gaussian process approximation and the usage of the formula~\eqref{gp_approx_variance} are not evident as one might directly estimate the variance of neural network prediction \(\hat{f}(\xv, \wv)\) at any point \(\xv\) by sampling from \(q(\wv)\) and use it as acquisition function. However, the approximate posterior variance \(\hat{\sigma}^2(\xv \cond \hat{f}, X)\) of Gaussian process \(\hat{g}(\xv \cond \hat{f})\) has an important property that is has large values for points \(\xv\) lying far from the points from the training set \(X\). Thus, out-of-sample points are likely to be selected by the active learning procedure.

  Moreover, the function \(\hat{\sigma}^2(\xv \cond \hat{f}, X)\) depends solely on covariance function values for points from a set \(X\) (and not on the output function values). Such property allows updating uncertainty predictions by just adding sample points to the set \(X\). More specifically, if we decide to sample some point \(\xv'\), then the updated posterior variance \(\hat{\sigma}^2(\xv \cond \hat{f}, X')\) for \(X' = X \cup \xv'\) can be easily computed:
  \begin{EQA}[c]
    \hat{\sigma}^2(\xv \cond \hat{f}, X') = \hat{\sigma}^2(\xv \cond \hat{f}, X) - \frac{\hat{k}^2(\xv, \xv' \cond \hat{f}, X)}{\hat{\sigma}^2(\xv' \cond \hat{f}, X)},
  \end{EQA}
  where \(\hat{k}(\xv, \xv' \cond \hat{f}, X) = \hat{k}(\xv, \xv') - \hat{\kv}^{\T}(\xv) \hat{K}^{-1} \hat{\kv}(\xv')\) is the posterior covariance function of the process \(\hat{g}(\xv \cond \hat{f})\) given \(X\).

\begin{figure*}[ht!]
    \centering
    \includegraphics[scale=.3]{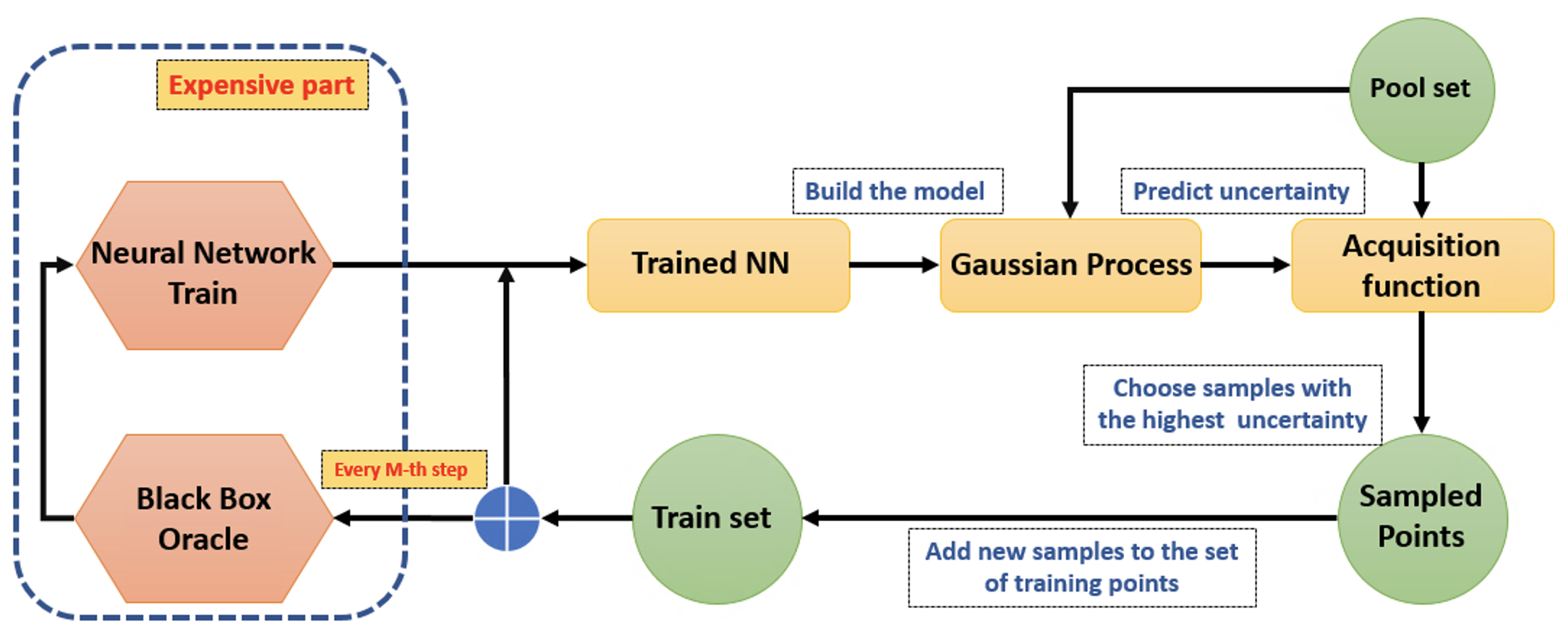} 
      \caption{Schematic representation of the NNGP approach to active learning. GP is fitted on the data from a stochastic output of NN, and the posterior variance of GP is used as an acquisition function for sampling. The most computationally expensive part (function evaluation at sampled points and neural network retraining) is done only every M steps of sampling, while all the intermediate iterations are based solely on trained neural network and corresponding GP approximation.}
    \label{fig:nngp-diag}
  \end{figure*}

  Importantly, \(\hat{\sigma}^2(\xv \cond \hat{f}, X')\) for points \(\xv\) in some vicinity of \(\xv'\) will have low values, which guarantees that further sampled points will not lie too close to \(\xv'\) and other points from the training set \(X\). The resulting NNGP active learning procedure is depicted in Figure~\ref{fig:nngp-diag}.

\section{Experiments}
  Our experimental study is focused on real-world data to ensure that the algorithms are indeed useful in real-world scenarios. We compare the following algorithms:
  \begin{itemize}
    \item pure random sampling;
    \item sampling based on the variance of NN stochastic output from $\hat{f}$, which we refer to as MCDUE (see~\cite{Gal2017,Tsymbalov2018});
    \item the proposed GP-based approaches (NNGP).
  \end{itemize}
  The more detailed descriptions of these methods can be found in Appendix~\ref{appendix:al_details}.

  Following~\cite{Gal2015} we use the Bayesian NN with the Bernoulli distribution on the weights, which is equivalent to using the dropout on the inference stage.

\subsection{Airline delays dataset and NCP comparison}
\label{sec:ncp}
  We start the experiments from comparing the proposed approach to active learning with the one based on uncertainty estimates obtained from a Bayesian neural network with \textit{Noise Contrastive Prior (NCP)}, see~\cite{Hafner2018}. Following this paper, we use the airline delays dataset (see~\cite{Hensman2013}) and NN consisting of two layers with 50 neurons each, leaky ReLU activation function, and trained with respect to NCP-based loss function.

  We took a random subset of 50,000 data samples from the data available on the January -- April of 2008 as a training set and we chose 100,000 random data samples from May of 2008 as a test set. We used the following variables as input features \textit{PlaneAge, Distance, CRSDepTime, AirTime, CRSArrTime, DayOfWeek, DayofMonth, Month} and \textit{ArrDelay + DepDelay} as a target. 
    
\begin{figure}[ht!]
    \centering
    \includegraphics[scale=.35]{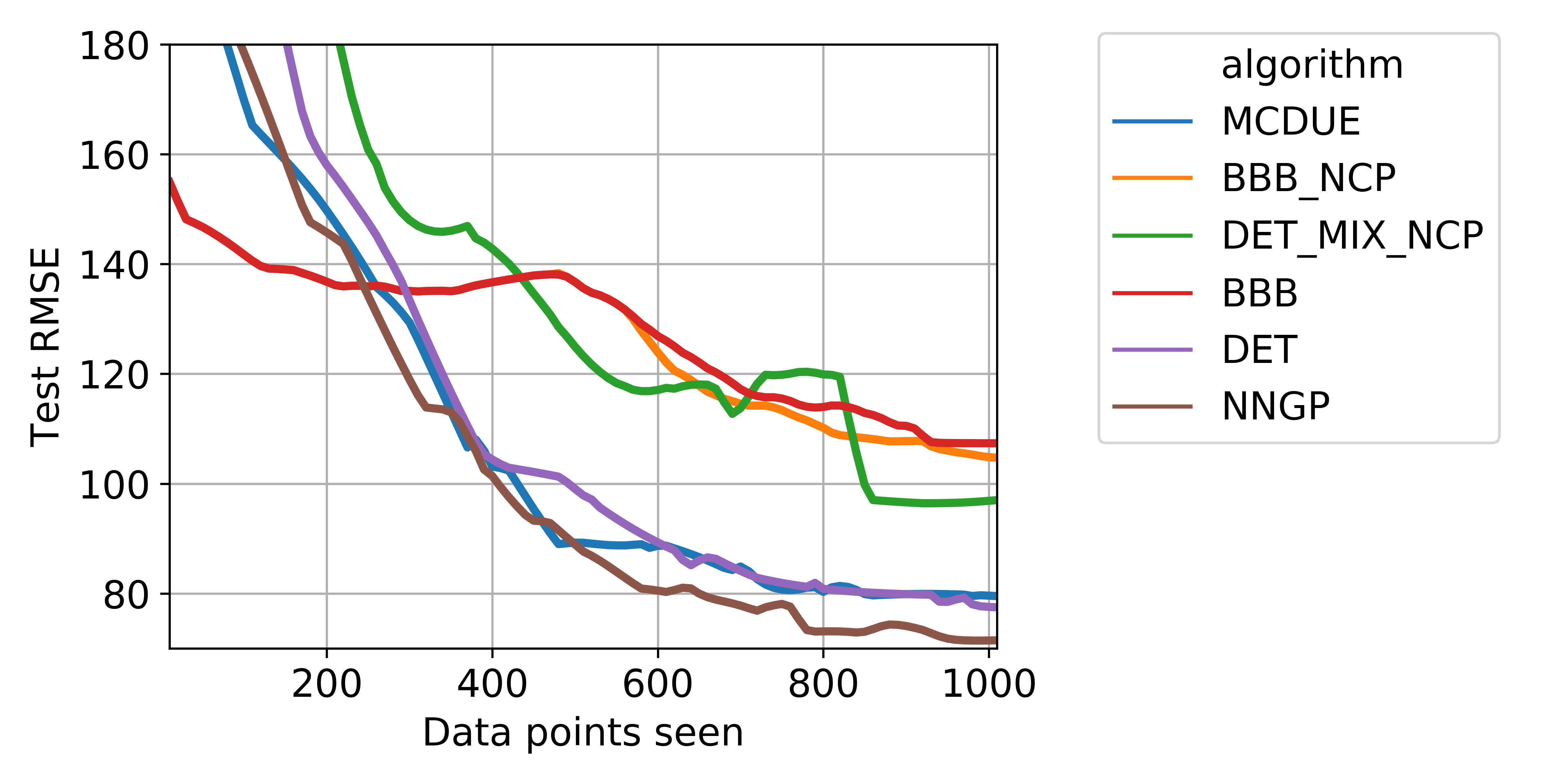} 
    \caption{Root mean squared errors as functions of active learning iteration for different methods on the Airline delays data set. Plots show median of the errors over 25 runs. NNGP initially has a much higher error, but shows the rapid improvement and becomes the best method near iteration 300.}
  \label{fig:flight-test}
  \end{figure}

  The results for the test set are shown in Figure~\ref{fig:flight-test}.
  The proposed NNGP approach demonstrates a comparable error with respect to the previous results and outperforms (on average) other methods in the continuous active learning scenario.

\subsection{Experiments on UCI datasets}
\label{sec:uci}
  We conducted a series of experiments with active learning performed on the data from the UCI ML repository~\cite{uci}. All the datasets represent real-world regression problems with \(15+\) dimensions and \(30000+\) samples, see Table~\ref{tab:datasets} for details. 

  \begin{table*}[ht!]
    \centering
    \caption{Summary of the datasets used in experiments with UCI data.}
    \label{tab:datasets}
    \scalebox{0.77}{
      \begin{tabular}{|c|c|c|c|c|c|c|}
        \hline
        \textbf{Dataset name} & \textbf{\# of samples} & \textbf{\# of attributes} & \textbf{Feature to predict} \\
        \hline
        BlogFeedback~\cite{buza2014} & 60021 & 281 & Number of comments \\
        \hline
        SGEMM GPU~\cite{nugteren2015} & 241600 & 18 & Median calculation time \\
        \hline
        YearPredictionMSD~\cite{bm2011} & 515345 & 90 & Year \\
        \hline
        Relative location of CT slices~\cite{graf2011} & 53500 & 386 & Relative location \\
        \hline
        Online News Popularity~\cite{fernandes2015} & 39797 & 61 & Number of shares \\ \hline
        KEGG Network~\cite{shannon2003cytoscape} & 53414 & 24 & Clustering coefficient \\ \hline
      \end{tabular}
    }
  \end{table*}

  For every experiment, data are shuffled and split in the following proportions: 10\% for the training set \(D_{train}\), 5\% for the test set \(D_{test}\), 5\% for the validation set \(D_{val}\) needed for early-stopping and 80\% for the pool \(\mathcal{P}\).

  %
  We used a simple neural network with three hidden layers of sizes 256, 128 and 128.
  The details on the neural network training procedure can be found in Appendix~\ref{appendix:training}.
  We performed \(16\) active learning iterations with 200 points picked at each iteration.


  To compare the performance of the algorithms across the different datasets and different choices of training samples, we will use so-called \textit{Dolan-More curves}. Let \(q^p_a\) be an error measure of the \(a\)-th algorithm on the \(\mathcal{P}\)-th problem. Then, determining the performance ratio \(r^p_a = \frac{q^p_a}{\min_x(q^p_x)}\), we can define the Dolan-More curve as a function of the performance ratio factor \(\tau\):
  \begin{EQA}[c]
    \rho_a(\tau) = \frac{\#(p\colon r^p_a \leq \tau)}{n_p},
  \end{EQA}
  where \(n_p\) is the total number of evaluations for the problem \(p\). Thus, \(\rho_a(\tau)\) defines the fraction of problems in which the \(a\)-th algorithm has the error not more than \(\tau\) times bigger than the best competitor in the chosen performance metric. Note that \(\rho_a(1)\) is the ratio of problems on which the \(a\)-th algorithm performance was the best, while in general, the higher curve means the better performance of the algorithm.

\begin{figure}[ht!]
    \hspace*{-1cm} 
    \centering
    \includegraphics[scale=.12]{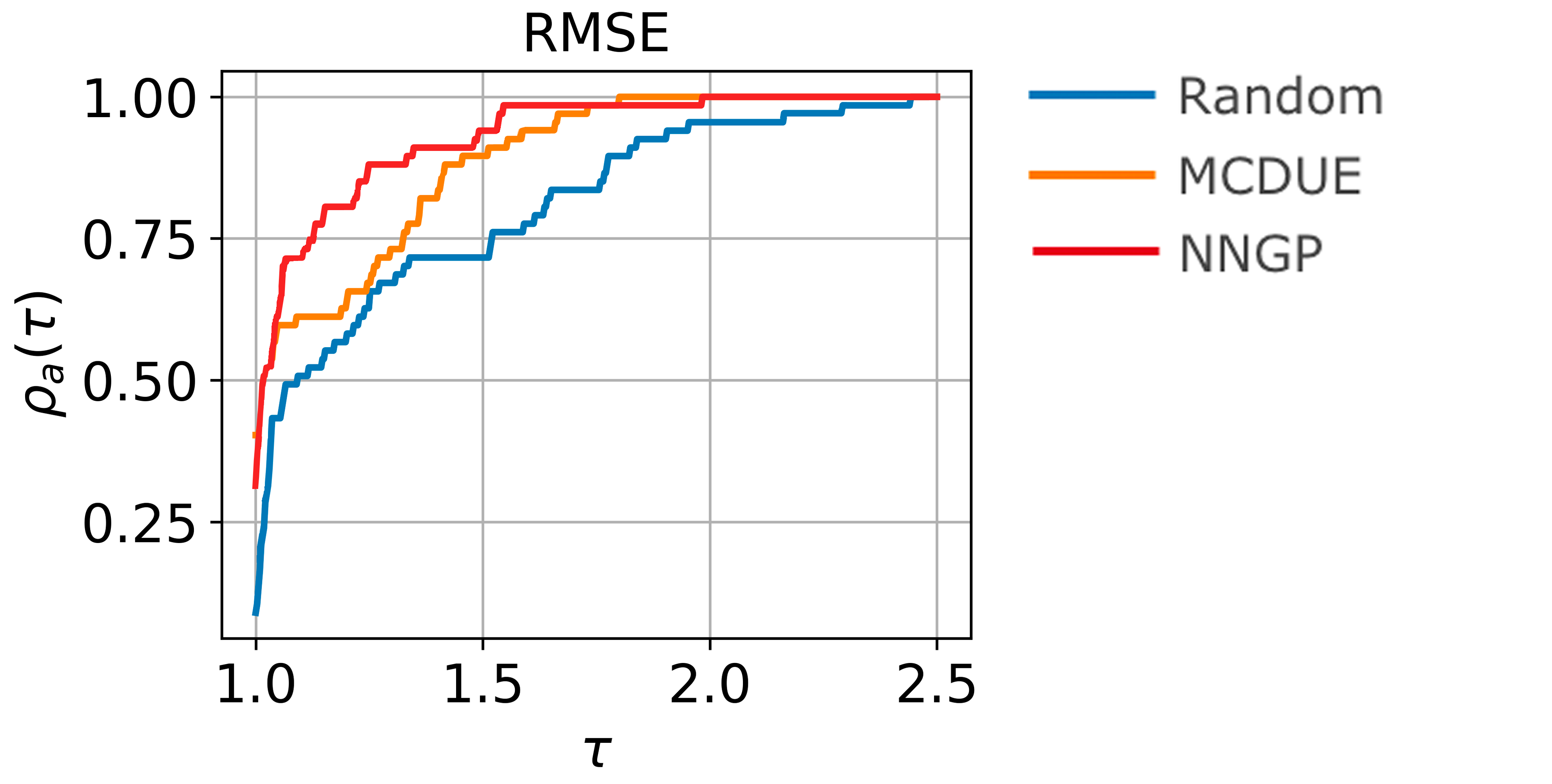} 
        \caption{Dolan-More curves for UCI datasets and different active learning algorithms after 16 active learning iterations. Root mean squared error (RMSE)
        on independent test set is considered.
        NNGP-based algorithm show better performance
        compared to MCDUE and random sampling.}
    \label{fig:dolan-more}
  \end{figure}

  The Dolan-More curves for the errors of approximation for considered problems after the 16th iteration of the active learning procedure are presented in Figure~\ref{fig:dolan-more}. We see that the NNGP procedure is superior in terms of RMSE compared to MCDUE and random sampling. 

\subsection{SchNet training}
  To demonstrate the power of our approach, we conducted a series of numerical experiments with the state-of-the-art neural network architecture in the field of chemoinformatics ``SchNet''~\cite{schutt}. This network takes information about an organic molecule as an input, and, after special preprocessing and complicated training procedure, outputs some properties of the molecule (like energy). Despite its complex structure, SchNet contains fully connected layers, so it is possible to use a dropout in between them.
    
  We tested our approach on the problem of predicting the internal energy of the molecule at 0K from the QM9 data set~\cite{ramakrishnan2014quantum}. We used a Tensorflow implementation of a SchNet with the same architecture as in original paper except for an increased size of hidden layers (from 64 and 32 units to 256 and 128 units, respectively) and dropout layer placed in between of them and turned on during an inference only.
    
  In our experiment, we separate the whole dataset of 133\,885 molecules into the initial set of 10\,000 molecules, the test set of 5\,000 molecules, and the rest of the data allocated as the pool. On each active learning iteration, we perform 100\,000 training epochs and then calculate the uncertainty estimates using either MCDUE or NNGP approach. We then select 2\,500 molecules with the highest uncertainty from the pool, add them to the training set and perform another active learning iteration.

\begin{figure}[ht!]
    \hspace*{-1cm} 
    \centering
    \includegraphics[scale=.255]{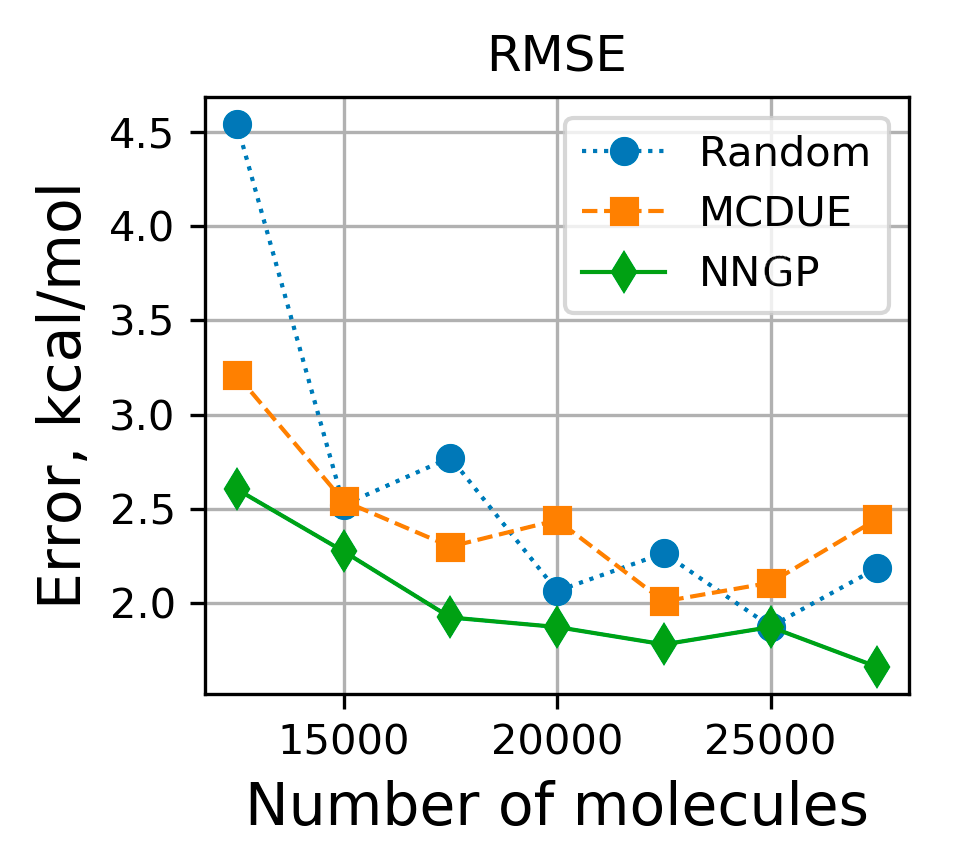} 
    \caption{Training curves for the active learning scenario for SchNet: starting from 10\,000 random molecules pick 2\,500 based on the uncertainty estimate. NNGP-based algorithm results in 25\% decrease in RMSE.
    Simple dropout-based approach (MCDUE) does not demonstrate a difference from the random sampling in terms of accuracy.}
  \label{fig:schnet-2}
  \end{figure}

  The results are shown in Figure~\ref{fig:schnet-2}. The NNGP approach demonstrates the most steady decrease in error, with the 25\% accuracy increase in RMSE.
  Such improvement is very significant in terms of the time savings for the computationally expensive quantum-mechanical calculations. For example, to reach the RMSE of 2 kcal/mol starting from the SchNet trained on 10\,000 molecules, one need to additionally sample 15\,000 molecules in case of random sampling or just 7\,500 molecules using the NNGP uncertainty estimation procedure.

    
    
    
    
\subsection{Hydraulic simulator}
  In the oil industry, to determine the optimal parameters and control the drilling process, engineers carry out hydraulic calculations of the well's circulation system, that usually are based on either empirical formulas or semi-analytical solutions of the hydrodynamic equations. However, such semi-analytical solutions are known just for few individual cases, while in the other ones, only very crude approximations are usually available (see~\cite{Podryabinkin2013} for details). As a result, such calculations have relatively low precision. On the other hand, the full-scale numerical solution of the hydrodynamic equations describing the flow of drilling fluids can provide a sufficient level of accuracy, but it requires significant computational resources and subsequently is very costly and time-consuming. The possible solution to this problem is the use of a surrogate model.

  We used a surrogate model for the fluid flow in the well-bore while drilling. The oracle, in this case, is a numerical solver for the hydrodynamic equations~\cite{Podryabinkin2013}, which, given six main parameters of the drilling process as an input, outputs the unit-less hydraulic resistance coefficient that characterizes the drop in the pressure.

\begin{figure*}[ht!]
    \hspace*{-1cm} 
    \centering
    \includegraphics[scale=.455]{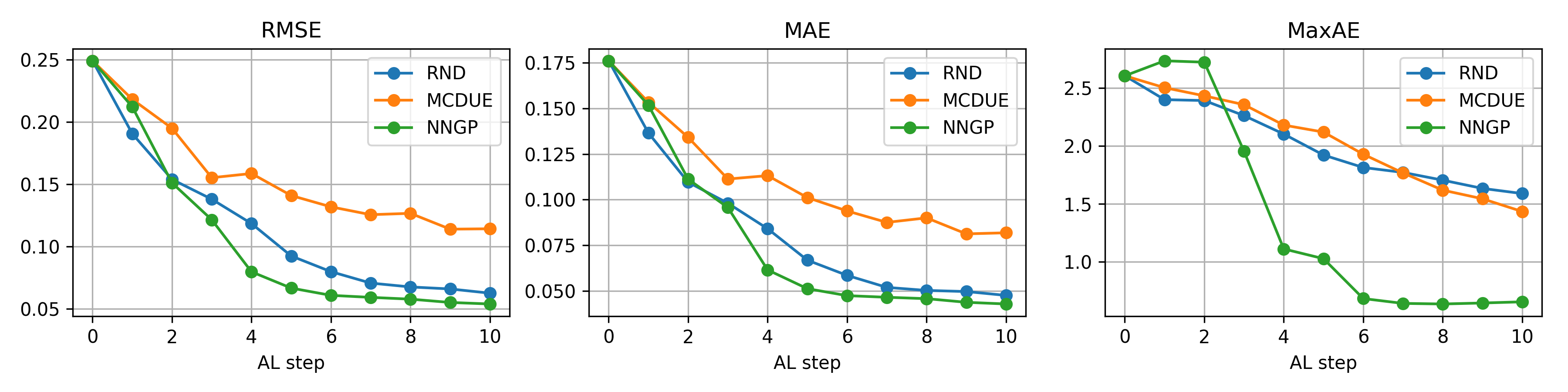} 
    \caption{Training curves for the active learning scenario for the hydraulic simulator case. The NNGP sampling algorithm outperforms the random sampling and MCDUE with a large margin in terms of maximal error.}
  \label{fig:hydraulic}
  \end{figure*}

  In this experiment, we used a two-layer neural network with \(50\) neurons per each layer and LeakyReLU activation function. Initial training and pool sets had \(50\) and \(20000\) points respectively. We completed \(10\) active learning iterations, adding \(50\) points per each iteration. The results are shown in Figure~\ref{fig:hydraulic}. Clearly, NNGP is superior in terms of RMSE and MAE, while maximum error for NNGP is 10 times lower.
  
\section{Related work}
\label{sec:related_work}

\subsection{Active learning}
  \textbf{Active learning}~\cite{Settles2012} (also known as \textit{adaptive design of experiments}~\cite{Forrester2008} in statistics and engineering design) is a framework which allows the additional data points to be annotated by computing target function value and then added to the training set. The particular points to sample are usually chosen as the ones that maximize so-called \textit{acquisition function}. The most popular approaches to construct acquisition function are committee-based methods~\cite{seung}, also known as \textit{query-by-committee}, where an ensemble of models is trained on the same or various parts of data, and Bayesian models, such as Gaussian Processes~\cite{Rasmussen2004}, where the uncertainty estimates can be directly inferred from the model~\cite{Sacks1989,Burnaev2015}. However, the majority of existing approaches are computationally expensive and sometimes even intractable for large sample sizes and input dimensions.
  
  \textbf{Active learning for neural networks} is not very well developed as neural networks usually excel in applications with large datasets already available. Ensembles of neural networks (see~\cite{Li2018} for a detailed review) often boil down to an independent training of several models, which works well in some applications~\cite{Beluch2018}, but is computationally expensive for the large-scale applications. Bayesian neural networks provide uncertainty estimates, which can be efficiently used for active learning procedures~\cite{Gal2017,Hafner2018}. The most popular uncertainty estimation approach for BNNs, \textit{MC-Dropout}, is based on classical dropout first proposed as a technique for a neural network regularization, which was recently interpreted as a method for approximate inference in Bayesian neural networks and shown to provide unbiased Monte-Carlo estimates predictive variance~\cite{Gal2015}. However, it was shown~\cite{Beluch2018} that uncertainty estimates based on MC-Dropout are usually less efficient for active learning then those based on ensembles. \cite{Pop2018} suggests that it is partially due to the mode collapse effect, which leads to overconfident out of sample predictions of uncertainty.

  \textbf{Connections between neural networks and Gaussian processes} recently gain significant attention, see~\cite{Matthews2018,Lee2017} which study random, untrained NNs and show that such networks can be approximated by Gaussian processes in infinite network width limit. Another direction is incorporation of the GP-like elements into the NN structure, see~\cite{Sun2018,Garnelo2018} for some recent contributions. However, we are not aware of any contributions making practical use of GP behaviour for standard neural network architectures.

\section{Summary and discussion}
    We have proposed a novel dropout-based method for the uncertainty estimation for deep neural networks, which uses the approximation of neural network by Gaussian process. Experiments on different architectures and real-world problems show that the proposed estimate allows to achieve state-of-the-art results in context of active learning. Importantly, the proposed approach works for any neural network architecture involving dropout (as well as other Bayesian neural networks), so it can be applied to very wide range of networks and problems without a need to change neural network architecture.
    
    It is of interest whether the proposed approach can be efficient for other applications areas, such as image classification.
    We also plan to study the applicability of modern methods for GP speed-up in order to improve the scalability of proposed approach.

\subsubsection*{Acknowledgements}
 The research was supported by the Skoltech NGP Program No. 2016-7/NGP (a Skoltech-MIT joint project).

\bibliographystyle{apalike}


\appendix

\section{Details on active learning approaches}
\label{appendix:al_details}

Here we provide descriptions for the uncertainty estimation and active learning algorithm used in experiments: MCDUE described in \cite{Gal2015,Tsymbalov2018} and NNGP / M-step NNGP proposed in this paper.

 \begin{algorithm}
 \label{algo: MCDUE}
    \caption{MCDUE}
    \begin{algorithmic}[1]
     \Require{Number of samples to generate \(N_{s}\), pool \(\pool\), neural network model \(\hat{f}(\xv, \wv)\) and dropout probability \(\pi\).}
     \Ensure{Set of points \(X_{s} \subset \pool\) with \(|X_{s}| = N_{s}\)}.
     \For{each sample \(\xv_j\) from the pool \(\mathcal{P}\)}
        \For{\(t = 1, \dots, T\)}
         \State \(\omega_t \sim \operatorname{Bern}(\pi)\).
         \State \(\wv_t = \hat{\wv}(\omega_t)\).
         \State \(y_t = \hat{f}_{t}(\xv_j) = \hat{f}(\xv_j, \wv_t)\).
        \EndFor

     \State Calculate the variance: 
        \begin{EQA}[c]
         \hat{\sigma}_j^2 = \frac{1}{T - 1} \sum_{t = 1}^T (y_t - \bar{y})^2,\ \bar{y} = \frac{1}{T} \sum_{t = 1}^T y_t.
        \end{EQA}
     \EndFor

     \State Return \(N_{s}\) points from pool \(\pool\) with largest values of the variance \(\hat{\sigma}_j^2\).
    \end{algorithmic}
 \end{algorithm}

 \begin{algorithm}
 \label{algo: NNGP}
    \caption{NNGP}
    \begin{algorithmic}[1]
     \Require{Number of samples to generate \(N_{s}\), pool \(\pool\), the set \(X_*\) of inducing points for Gaussian process model, neural network model \(\hat{f}(\xv, \wv)\), dropout probability \(\pi\) and regularization parameter \(\lambda\).}
     \Ensure{Set of points \(X_{s} \subset \pool\) with \(|X_{s}| = N_{s}\)}.

     \For{\(t = 1, \dots, T\)}
        \State \(\omega_t \sim \operatorname{Bern}(\pi)\).
        \State \(\wv_t = \hat{\wv}(\omega_t)\).
        \State \(y^i_t = \hat{f}(\xv_i, \wv_t)\) for each \(\xv_i \in \mathcal{P}\).
        \State \(z^j_t = \hat{f}(\xv_j, \wv_t)\) for each \(\xv_j \in X_*\).
     \EndFor

     \State Calculate the covariance matrix \(\hat{K} = \bigl[cov(z^i, z^j)\bigr]_{i, j = 1}^{N}\).

     \For{each \(\xv_j \in \mathcal{P}\)}
        \State \(\hat{\kv}_j = \bigl[cov(z^i, y^j)\bigr]_{i = 1}^{N}\).
        \State \(v_j = var(y^j)\).
        \State \(\hat{\sigma}_j^2 = v_j - \hat{\kv}_j^{\T} (\hat{K} + \lambda I)^{-1} \hat{\kv}_j\).
     \EndFor

     \State Return \(N_{s}\) points from pool \(\pool\) with largest values of the variance \(\hat{\sigma}_j^2\).
    \end{algorithmic}
 \end{algorithm}

 \begin{algorithm}
 \label{algo: MNNGP}
    \caption{M-step NNGP}
    \begin{algorithmic}[1]
     \Require{Number of samples to generate \(N_{s}\),  number of samples per active learning iteration \(M\), pool \(\pool\), the set \(X_*\) of inducing points for Gaussian process model, neural network model \(\hat{f}(\xv, \wv)\), dropout probability \(\pi\) and regularization parameter \(\lambda\).}
     \Ensure{Set of points \(X_{s} \subset \pool\) with \(|X_{s}| = N_{s}\)}.
     
     \State Initialize sets \(X_{s} := \emptyset, \pool^* := \mathcal{P}\).

     \For{\(m = 1, \dots, M\)}
        \State Run NNGP procedure with parameters \(N_s / M, \pool^*\), \(X_*~\cup~X_{s}, \hat{f}, \pi, \lambda\), which returns a set of points \(X_o\).

        \State \(X_s = X_s \cup X_o\).
        \State \(\pool^* = \pool^* \setminus X_{o}\).
     \EndFor
    \end{algorithmic}
 \end{algorithm}

\section{NN training details}
\label{appendix:training}

 This section provides details on NN training for UCI dataset experiments.
 
 Neural network had three hidden layers with sizes 256, 128, 128, respectively. Learning rate started at \(10^{-3}\), its decay was set to 0.97 and changed every 50000 epochs. Minimal learning rate was set to \(10^{-5}\). We reset learning rate for each active learning algorithm in hope of beating the local minima problem. Training dropout rate was set to 0.1. \(L_2\) regularization was set to \(10^{-4}\). Batch size set to 200.

 We developed rather complex active learning procedure in order to balance between the stochastic nature of NN training and early stopping.

 The algorithm we followed for each experiment on a fixed dataset to initialize the training is as follows:
 \begin{enumerate}
    \item Initialize NN. Shuffle and split the data. Train for a mandatory number of epochs \(epochs_{mandatory} = 10000\). Set \(warnings = 0\), \(E^{previous}_{val} = 10^{10}\).
    
    \item For every epoch \(current\_epoch\) in \(1, \ldots, epochs_{max} = 10^6\):
    \begin{enumerate}
     \item Train NN.
    
     \item If \(current\_epoch\) \% \(step_{ES\_check} = 100\):
        \begin{enumerate}
         \item Get RMSE error \(E_{val}\) on validation set.
    
         \item If current \(E_{val}\) exceeds the \(E^{previous}_{val}\) by \(ES_{window} = 1\%\): then \(warnings := warnings + 1\). Else set \(warnings := 0, E^{previous}_{val} := E_{val}\).
    
         \item If \(warnings > warnings_{max} = 3\): break from training procedure.
        \end{enumerate}
    \end{enumerate}
 \end{enumerate}

\end{document}